%% file: main.tex
\documentclass[10pt,twocolumn,letterpaper]{article}

\usepackage[pagenumbers]{cvpr} 

\input{preamble}
\usepackage{colortbl}
\usepackage{fontawesome}
\usepackage{makecell}
\usepackage{algorithm}
\usepackage{algorithmic}     
\usepackage{graphicx}
\usepackage{adjustbox}
\usepackage{tcolorbox}
\tcbuselibrary{listings,breakable} 
\definecolor{cvprblue}{rgb}{0.21,0.49,0.74}
\definecolor{cyan}{RGB}{230,240,255}
\definecolor{pink}{RGB}{255,192,203}
\definecolor{ifedit}{HTML}{008B8B}
\definecolor{ifeditlight}{RGB}{200,240,240}
\usepackage[pagebackref,breaklinks,colorlinks,allcolors=cvprblue]{hyperref}

\def\paperID{1397} 
\def\confName{CVPR}
\def\confYear{2026}

\title{Are Image-to-Video Models Good Zero-Shot Image Editors?}

\author{%
  Zechuan Zhang$^{1}$~~ Zhenyuan Chen$^{1}$~~ Zongxin Yang$^{2}$~~ Yi Yang$^{1 \dag}$ \\
  $^1$~Zhejiang University  ~~ $^2$~Harvard University
}

\begin{document}

\input{figure_latex/teaser}
\maketitle
\input{sec/0_abstract}

\input{sec/1_intro}

\input{sec/2_Related}
\input{sec/2z_Preliminary}

\input{sec/3_Method}
\input{sec/4_Experiment}
\input{sec/5_Conclusion}

\clearpage
{
    \small
    \bibliographystyle{ieeenat_fullname}
    \bibliography{main}
}

\end{document}

%% file: preamble.tex
\newcommand{\ifedit}[1]{\textcolor[HTML]{008B8B}{#1}}



%% file: figure_latex/teaser.tex
\newcommand{\teaserCaption}{
\textbf{Visual results of \ifedit{IF-Edit}.} 
We propose \ifedit{IF-Edit} (\textbf{I}mage Editing by Generating \textbf{F}rames), a tuning-free framework that repurposes image-to-video diffusion models for zero-shot image editing. 
By leveraging the world-simulation priors of video models and our proposed modules, \ifedit{IF-Edit} achieves physically consistent and semantically aligned edits, excelling in \textbf{non-rigid transformations}, \textbf{temporal progression}, and \textbf{causal reasoning} scenarios.
}

\twocolumn[{
    \renewcommand\twocolumn[1][]{#1}
    \maketitle
    \centering
    \begin{minipage}{1.00\textwidth}
        \centering 
        \includegraphics[width=\linewidth]{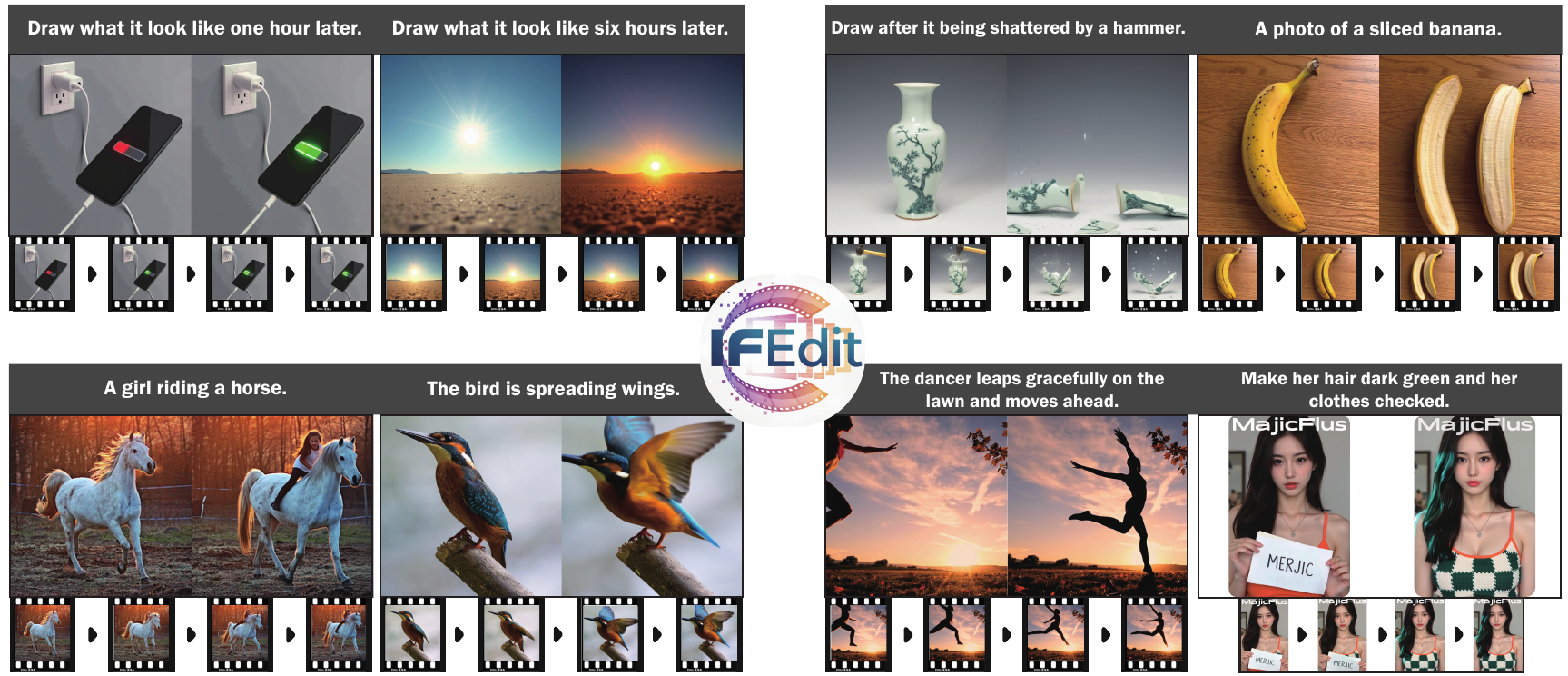}
    \end{minipage}
      \vspace{-0.5 em}
    \captionsetup{type=figure}
    \captionof{figure}{\looseness=-1{\teaserCaption}}
    \label{fig:teaser}
    \vspace{1.5 em}
}]

%% file: sec/0_abstract.tex
\begin{abstract}
Large-scale video diffusion models exhibit strong world-simulation and temporal reasoning capabilities, yet their potential as zero-shot image editors remains underexplored. 
We present \ifedit{IF-Edit} (\textbf{I}mage Edit by Generating \textbf{F}rames), a tuning-free framework that repurposes pre-trained image-to-video diffusion models for instruction-driven image editing. 
\ifedit{IF-Edit} addresses three core obstacles—prompt misalignment, redundant temporal latents, and blurry late-stage frames—via: (1) a Chain-of-Thought Prompt Enhancement module that reformulates static editing instructions into temporally grounded reasoning prompts; (2) a Temporal Latent Dropout strategy that compresses frame latents after the expert-switch point, accelerating denoising while preserving global semantics and temporal coherence; and (3) a Self-Consistent Post-Refinement step that refines the sharpest late-stage frame through a brief still-video trajectory, leveraging the video prior for sharper and more faithful results. 
Extensive experiments across four public benchmarks—covering non-rigid deformations, physical and temporal reasoning, and general instruction editing—show that \ifedit{IF-Edit} achieves strong performance on non-rigid and reasoning-centric tasks while remaining competitive on general-purpose edits. 
Our study offers a systematic view of video diffusion models as image editors, revealing their unique strengths, limitations, and a simple recipe for unified video–image generative reasoning.
\end{abstract}

%% file: sec/1_intro.tex
\section{Introduction}
\label{sec:intro}

Recent advances in large-scale generative models have transformed image editing from creative retouching into a versatile tool for simulation, education, and content creation~\cite{wu2025chronoedittemporalreasoningimage,zhuSceneCrafterControllableMultiView2025,zengEditWorldSimulatingWorld2025,bianEffectsAIgeneratedImages2025,weiEffectsGenerativeAI2025,yu2025anyedit,jiInstructionbasedImageEditing2025,zhou2024migc,zhou20243dis,xu2025contextgen}. 
Modern systems such as GPT-Image~\cite{gpt4o} and Nano Banana~\cite{google2025gemini25flashmodelcard} allow users to describe edits in natural language and obtain high-quality results, supporting applications in social media, design, and e-commerce. 
Beyond these cases, controllable editing also supports simulation and safety-critical tasks that require synthesizing rare or hard-to-capture conditions.
However, most existing systems operate on a \emph{single-frame} representation and lack explicit temporal or causal priors that could support more complex, reasoning-driven transformations.

Most text-guided image editing methods formulate the task as image-to-image translation. 
Approaches such as~\cite{brooks2023instructpix2pix,zhang2023magicbrush,zhao2024ultraedit,zhang2025context,liu2025step1x,yu2025anyedit,dengEmergingPropertiesUnified2025} excel at localized edits—adding, removing, or replacing attributes—but struggle with large viewpoint changes, long-range physical reasoning, or maintaining self-consistency under substantial geometric transformations. 
Recent multimodal systems~\cite{gpt4o,google2025gemini25flashmodelcard,wu2025qwen,xie2025reconstruction} can handle certain reasoning-based edits, yet rely on costly finetuning over curated datasets. 
Datasets~\cite{qian2025picobanana400klargescaledatasettextguided,Sheynin_2024_CVPR,chenOpenGPT4oImageComprehensive2025,wangGPTIMAGEEDIT15MMillionScaleGPTGenerated2025a,maX2EditRevisitingArbitraryInstruction2025} require thousands of paired samples to ensure text–image alignment, further increasing the computational barrier. 
Overall, while these methods achieve impressive visual quality, they remain confined to single-frame modeling and do not explicitly exploit temporal or causal structure.


Recent progress in large-scale video diffusion models~\cite{yang2024cogvideox,wan2p1,pika,wiedemer2025videomodelszeroshotlearners,maLatteLatentDiffusion2025} has revealed remarkable world-simulation capabilities: these models produce smooth, temporally coherent videos that follow realistic physics and maintain object consistency over time—behaviors reminiscent of \emph{world models}~\cite{ha2018world} that simulate plausible dynamics rather than isolated images.  
Recent studies~\cite{wiedemer2025videomodelszeroshotlearners,wu2025chronoedittemporalreasoningimage} further show that video diffusion models exhibit \emph{chain-of-frames} reasoning—an analogue to chain-of-thought in language models—when generating short clips from a single static image as the first frame, effectively performing step-by-step transformations guided by text instructions. 
These observations raise a natural question: \textbf{Can image-to-video diffusion models be directly used as tuning-free, general-purpose image editors by leveraging their temporal priors, without any additional finetuning?}

\input{figure_latex/high_level_compare}

While several recent works~\cite{rotstein2025pathwaysimagemanifoldimage,wiedemer2025videomodelszeroshotlearners,wu2025chronoedittemporalreasoningimage} explore adapting video diffusion models for image editing, current approaches still face three major limitations (\cref{fig:high level compare}). 
(1) \textbf{Redundant computation.} 
Video generators typically produce dozens of frames, whereas image editing ultimately requires a single output frame, leading to substantial wasted computation. Methods such as ChronoEdit~\cite{wu2025chronoedittemporalreasoningimage} reduce this cost via finetuning, but at the price of large-scale training. 
(2) \textbf{Inefficient frame selection.} 
Because many frames may satisfy the editing prompt, existing works often rely on vision–language models (VLMs) or manual inspection to select the final result~\cite{rotstein2025pathwaysimagemanifoldimage,wiedemer2025videomodelszeroshotlearners}, which improves accuracy but introduces latency and engineering complexity. 
(3) \textbf{Limited systematic understanding.} 
Despite rapid progress, there is still little systematic analysis of how off-the-shelf image-to-video models behave as zero-shot image editors across both general and reasoning-centric benchmarks.

In this work, we present \ifedit{IF-Edit} (\textbf{I}mage Edit by Generating \textbf{F}rames), a tuning-free and efficient framework that repurposes pre-trained image-to-video diffusion models for instruction-driven image editing. 
Rather than designing a new editor, \ifedit{IF-Edit} revisits the editing pipeline and addresses three core obstacles—prompt misalignment, redundant temporal latents, and blurry late-stage frames—through three lightweight components. 
(1) A \textbf{Chain-of-Thought Prompt Enhancement} module leverages a vision–language model to jointly interpret the input image and instruction, converting them into temporally grounded, reasoning-aware prompts that better match the video model’s world-simulation prior. 
(2) A \textbf{Temporal Latent Dropout} strategy removes redundant temporal latents after the early denoising stage has established the global layout, retaining only key frames to preserve temporal coherence while greatly reducing computation. 
(3) A \textbf{Self-Consistent Post-Refinement} step identifies the sharpest late-stage frame and performs a brief still-video refinement using the same model, yielding clearer textures and more faithful details through self-aligned enhancement.

To assess video models as image editors, we evaluate \ifedit{IF-Edit} across four public benchmarks spanning three dimensions: 
\textbf{non-rigid/motion} (TEdBench~\cite{Kawar_2023_CVPR}, ByteMorph~\cite{chang2025bytemorphbenchmarkinginstructionguidedimage}), 
\textbf{reasoning} (RISEBench~\cite{zhao2025envisioningpixelsbenchmarkingreasoninginformed}), 
and \textbf{general editing} (ImgEdit~\cite{ye2025imgedit}). 
Experiments show that \ifedit{IF-Edit} achieves strong performance on non-rigid and reasoning-centric tasks while remaining competitive with strong open-source systems on general instruction editing, highlighting the unique advantages of video priors for temporally and physically coherent edits. 
At the same time, performance on generic attribute/style changes lags behind specialized image editors, revealing clear room for improvement. 
Overall, our study provides a systematic view of off-the-shelf image-to-video models as image editors, clarifying both their strengths and their limitations.

In summary, our contributions are three-fold:
\begin{itemize}
    \item We introduce \ifedit{IF-Edit}, a tuning-free framework that repurposes pre-trained image-to-video diffusion models for zero-shot, instruction-driven image editing without any additional finetuning.
    \item We propose three lightweight components—VLM-guided chain-of-thought prompt enhancement, temporally aligned latent compression via \textbf{Temporal Latent Dropout (TLD)}, and \textbf{Self-Consistent Post-Refinement (SCPR)}—that jointly reduce redundancy and improve temporal and perceptual consistency.
    \item We conduct a systematic empirical study of off-the-shelf image-to-video models on diverse image-editing benchmarks, revealing where video priors excel (non-rigid and reasoning-centric edits) and where they still fall short (general instruction editing).
\end{itemize}

%% file: figure_latex/high_level_compare.tex
\begin{figure}[t]
\centering
\scriptsize
\includegraphics[width=\linewidth]{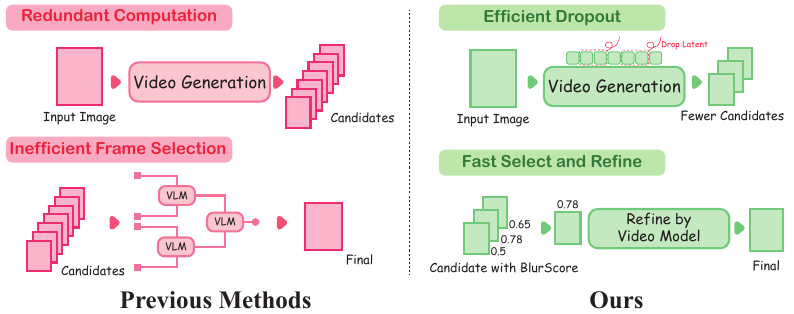}  
 \vspace{-2.0 em}
\scriptsize
\caption{
\textbf{Comparison with previous methods.} 
Unlike prior approaches that suffer from redundant video generation and costly VLM-based frame selection, 
we design efficient strategies in both stages: an \textbf{efficient temporal dropout} to reduce redundant computation during generation, 
and a \textbf{fast self-consistent refinement} for sharp and high-quality final results.
}
\vspace{-2.0 em}
\label{fig:high level compare}
\end{figure} 

%% file: sec/2_Related.tex
\section{Related Work}
\label{sec:related}
\noindent\textbf{Image Editing} is defined as modifying the appearance, structure, or content of an existing image to achieve a desired outcome, most commonly guided by user-specified textual prompts~\cite{huangDiffusionModelBasedImage2025}. Image editing models are typically built upon pre-trained text-to-image diffusion models and face a fundamental dilemma: (1) training-free methods~\cite{routSemanticImageInversion2025,Mokady_2023_CVPR,mengSDEditGuidedImage2022,couaironDiffEditDiffusionbasedSemantic2023,xuInversionFreeImageEditing2024,Brack_2024_CVPR,parmarZeroshotImagetoImageTranslation2023} employ inversion~\cite{songDenoisingDiffusionImplicit2020,juPnPInversionBoosting2024} and attention manipulation~\cite{hertzPrompttoPromptImageEditing2022}, using source and target prompts to guide edits, which can yield effective results for object-level editing but often fail to capture complex semantic relationships; (2) in contrast, training-based methods~\cite{bar-talText2LIVETextDrivenLayered2022, brooks2023instructpix2pix, Huang_2024_CVPR, Kawar_2023_CVPR, zhang2023magicbrush, zhao2024ultraedit,yu2025anyedit} fine-tune pre-trained models, achieving superior performance at the cost of requiring large-scale paired datasets and substantial computational resources. Recently, thanks to strong text-to-image diffusion base models~\cite{FLUX,esserScalingRectifiedFlow2024} available, promising works investigate efficient adaption~\cite{zhang2025context} even training-free methods~\cite{avrahamiStableFlowVital2025,kulikovFlowEditInversionFreeTextBased2025,wangTamingRectifiedFlow2025,yoon2025splitflow,Xu_2025_CVPR,yang2025texttoimage} to build image editing models upon it. In addition, recent work has built strong image editing models using heavy computation budget in both diffusion manners ~\cite{caiHiDreamI1HighEfficientImage2025,liu2025step1x,wang2025ovisu1technicalreport,wu2025qwen} and auto-regressive ones~\cite{xiao2025omnigen,wu2025omnigen2,linUniWorldV1HighResolutionSemantic2025,li2025uniworldv2,dengEmergingPropertiesUnified2025,cuiEmu35NativeMultimodal2025,chen2025blip3onext}.

\noindent\textbf{Video Priors for Image Editing.}
Given the rapid progress of video generation models~\cite{sora,veo,veo2,veo3,wan2p1,gen2,gen3,svd,kongHunyuanVideoSystematicFramework2025a,pengOpenSora20Training2025a,zhengOpenSoraDemocratizingEfficient2024,dongWanAlphaHighQualityTexttoVideo2025,nvidia2025worldsimulationvideofoundation}, recent works have begun exploring how video priors can benefit image editing.  
Frame2Frame (F2F)~\cite{rotstein2025pathwaysimagemanifoldimage}, a representative training-free approach, uses pretrained video models but depends heavily on VLM-based post-selection to choose the best frame from long generated clips—introducing substantial latency.  
ChronoEdit~\cite{wu2025chronoedittemporalreasoningimage} takes the opposite route by finetuning video models to shorten trajectories and improve edit accuracy, but at the cost of large-scale training. In contrast, \textbf{IF-Edit occupies a unique middle ground}: fully training-free like F2F but without heavy VLM filtering, and competitive with ChronoEdit while requiring no finetuning.  
This reveals an efficient new path for leveraging video priors to build strong, temporally grounded image editors.

\noindent\textbf{World Models from a Single Image.}
A growing body of research is exploring video models as world simulators which are capable of generating dynamic, interactive worlds from a single image~\cite{guiImageWorldGenerating2025,gillmanForcePromptingVideo2025, liang2025wonderland,yu2025wonderworldinteractive3dscene,li20244k4dgen,wang2025vistadream,ma2025you,xiao2025worldmem}. In a similar spirit, \ifedit{IF-Edit} leverages a short, deterministic simulation to achieve a specific final state as defined by an editing instruction. Our Chain-of-Frames reasoning can be seen as a controlled, task-oriented traversal through this micro-world's state space, efficiently harnessing the model's simulation capabilities for a targeted editing task.

\vspace{-2 pt}

%% file: sec/2z_Preliminary.tex
\input{figure_latex/highlownoise_observe}
\input{figure_latex/method_pipeline}
\section{Preliminary}

\textbf{Image-to-Video Diffusion Models.} 
Given an input image $x_0$ and a text instruction $c$, image-to-video diffusion models synthesize a sequence of frames $\{x_t\}_{t=1}^F$ by iteratively denoising latent conditioned on both $x_0$ and $c$. 
To construct the conditioning latent, the image $x_0$ is first concatenated with $(F{-}1)$ zero-frame placeholders along the temporal dimension to form a pseudo-video $X=[x_0, \mathbf{0}_{1:F-1}]$, which is then encoded by a 3D VAE encoder~$\mathcal{E}$~\cite{svd,wuImprovedVideoVAE2025,yang2024cogvideox} into $Y=\mathcal{E}(X)$. 
A binary mask $m$ indicates the observed frame ($m_0=1$) and the remaining unobserved ones ($m_t=0$ for $t>0$). 
During generation, Gaussian noise latents $\{z_t\}_{t=1}^{F}$ are sampled and concatenated with $Y$ and $m$ along the channel dimension as input to the DiT backbone $\epsilon_\theta$: 
\vspace{-0.4\baselineskip}
\[
z_{t-1} = \epsilon_\theta([z_t, Y, m], t, c),
\]
allowing temporal information to propagate from the image and produce coherent frame transitions guided by prompt.

\noindent \textbf{Wan 2.2 Model.} 
To assess the upper bound of zero-shot editing with video diffusion, we adopt Wan 2.2~\cite{wan2p1}, the strongest open-source image-to-video model. 
Wan 2.2 employs a Mixture-of-Experts (MoE)~\cite{moe1,moe2} diffusion backbone with two denoising experts: a high-noise expert for early global layout and a low-noise expert for later detail refinement. 
Expert switching is governed by the signal-to-noise ratio $SNR(t)=\alpha_t^2/\sigma_t^2$, transitioning once it exceeds half of its minimum. 
Each expert has about 14B parameters (27B total, 14B active per step), enabling high-quality generation with nearly constant inference cost.

%% file: figure_latex/highlownoise_observe.tex
\begin{figure}[t]
\centering
\scriptsize
\includegraphics[width=\linewidth]{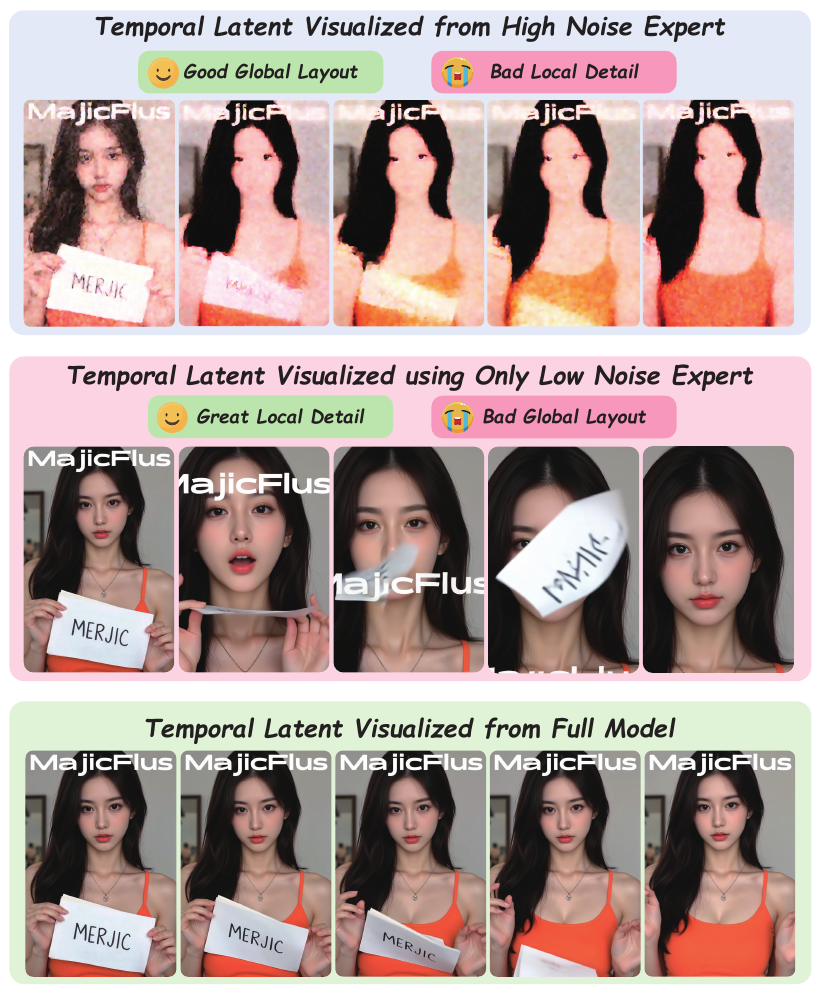}  
 \vspace{-2.0 em}
\scriptsize
\caption{\textbf{Visualization of expert behaviors. (\S\ref{sec:Temporal latent dropout})} 
Given the prompt “\textit{She loosens her grip, the card drifts down},” 
the \textbf{high-noise expert} quickly builds a coherent global layout but lacks fine detail, 
whereas the \textbf{low-noise expert} enhances local fidelity while losing spatial consistency. 
This observation motivates preserving early layout formation and refining details efficiently.}
\vspace{-2.0 em}
\label{fig:high low noise observation}
\end{figure} 

%% file: figure_latex/method_pipeline.tex
\newcommand{\pipelineCaption}{
\textbf{Overview of \ifedit{IF-Edit}.} (\S\ref{sec:method}) 
Our framework adapts an image-to-video diffusion model for zero-shot image editing through three components: 
(1) \textbf{Prompt Enhancement via CoT}, which reformulates static instructions into temporally grounded reasoning prompts; 
(2) \textbf{Temporal Latent Dropout (TLD)}, which accelerates inference by sparsifying temporal latents while preserving motion consistency; and 
(3) \textbf{Self-Consistent Post-Refinement}, which selects the sharpest frame via Laplacian score and performs still-video refinement to enhance detail and stability. 
Together, these modules enable efficient, physically consistent, and instruction-aligned image editing.
}

\begin{figure*}[th]
    \centering
    \scriptsize
    \includegraphics[width=\linewidth]{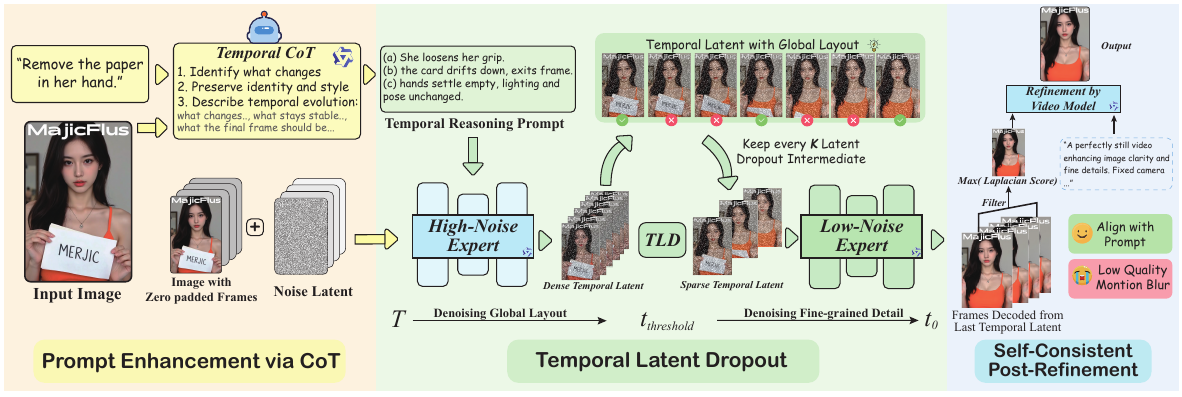}
    \vspace{-1.5 em}
    \captionof{figure}{\pipelineCaption}
     \vspace{-1.5 em}
    \label{fig:pipeline}
\end{figure*}

%% file: sec/3_Method.tex
\section{Method}\label{sec:method}

We propose \ifedit{IF-Edit}, a tuning-free framework that repurposes image-to-video diffusion models for zero-shot image editing.  
It harnesses the models’ intrinsic temporal reasoning while eliminating redundant video generation.  
As shown in~\cref{fig:pipeline}, our approach comprises three lightweight components:  
a \textbf{Prompt Enhancement via Chain-of-Thought} that reformulates static editing instructions into temporally grounded, process-aware prompts~(\cref{sec:prompt enhance via cot});  
a \textbf{Temporal Latent Dropout (TLD)} that accelerates denoising by sparsifying redundant temporal latents~(\cref{sec:Temporal latent dropout});  
and a \textbf{Self-Consistent Post-Refinement (SCPR)} that improves visual fidelity through self-aligned, post-diffusion enhancement using the same video prior~(\cref{sec:self-consistent post-refinement}).

\subsection{Prompt Enhancement via Chain-of-Thought}\label{sec:prompt enhance via cot}

Video diffusion models are trained to follow temporally grounded captions, yet standard editing instructions are often static and ambiguous. 
To bridge this gap, we employ a vision-language model (Qwen3-VL-A3B) to jointly interpret the input image and text instruction, transforming them into a \emph{chain-of-thought visual reasoning prompt} that describes how the scene evolves over time. 

Rather than directly rewriting the caption~\cite{rotstein2025pathwaysimagemanifoldimage}, 
the model infers a plausible temporal sequence—how elements move, appear, or transform—while preserving scene identity and visual continuity. 
This chain-of-thought reasoning~\cite{wei2022chain} \emph{not only helps the model correctly interpret user instructions but also externalizes its reasoning process, 
which benefits the video model’s temporal generation.} 
The resulting progression clearly encodes how the scene moves and changes over time, allowing the video diffusion model to treat editing as a smooth visual evolution rather than a single-frame modification.
The full system prompt and example outputs used for CoT rewriting are provided in the Appendix.

\subsection{Temporal Latent Dropout Strategy}\label{sec:Temporal latent dropout}

Video models exhibit strong world-simulation ability, generating temporally coherent frame sequences that follow physical dynamics and maintain object consistency. 
However, in image editing, only the \emph{final frame} matters, making full video generation computationally redundant. 
A strategy that retains temporal reasoning while reducing denoising cost is thus essential for practical efficiency.

We analyze the state-of-the-art open-source model Wan 2.2 and \textbf{observe distinct roles of its two denoising experts.} 
In early steps ($t>t_{\text{thresh}}$), the high-noise expert rapidly establishes global layout and coarse temporal structure, while in later steps ($t<t_{\text{thresh}}$) the low-noise expert refines texture and details. 
As shown in~\cref{fig:high low noise observation}, decoding early latents yields temporally coherent but blurry frames, while relying solely on the low-noise expert produces sharper yet semantically inconsistent results.  

Furthermore, we observe that if all intermediate frames are discarded and only the final latent is used for decoding, both image–input consistency and visual clarity degrade significantly.  
In contrast, retaining the first latent and a few intermediate ones preserves global consistency and fine details, as illustrated in the last row of~\cref{fig:high low noise observation}.  
These observations suggest that \textbf{temporal reasoning and global layout formation primarily occur in the early denoising stage}, and most intermediate temporal latents are redundant and can be safely dropped with minimal impact on final quality. We also measure this quantitatively in~\cref{tab:ablation}.

Based on this insight, we introduce a \textbf{Temporal Latent Dropout (TLD)} strategy. 
Let the latent at step $t$ be $z_t\in\mathbb{R}^{C\times F\times H\times W}$, where $F$ is the temporal length. 
Once the process reaches the regime switch to the low-noise expert ($t \le T_{\text{th}}$), 
we perform a \emph{one-shot} temporal subsampling:
\[
\tilde{z}_t = \mathcal{D}_K(z_t) = z_t[:,\, \{0,K,2K,\dots,F-1\},:,:],
\]
where $\mathcal{D}_K(\cdot)$ denotes the dropout operator. 
The reduced latent $\tilde{z}_t$ is then passed to the next denoising iteration. 
This operation effectively removes redundant temporal tokens while maintaining those necessary for temporal reasoning and the final result synthesis. 
We empirically set $K{=}3$ and $T_{\text{th}}$ near the expert-switch point.

If the original denoising cost scales with $\mathcal{O}(F)$ in the temporal dimension, TLD reduces it to roughly $\mathcal{O}(F/K)$ while maintaining comparable semantic consistency.  
Although introduced with Wan~2.2’s MoE design, TLD generalizes to other video diffusion models by adjusting the threshold $t_{\text{thresh}}$, since early denoising similarly captures global temporal structure. 
Generalization results on other video models are included in the Appendix.

\subsection{Self-Consistent Post-Refinement}\label{sec:self-consistent post-refinement}
Existing methods~\cite{rotstein2025pathwaysimagemanifoldimage,wiedemer2025videomodelszeroshotlearners} typically rely on vision-language models (VLMs) to score and select the best frame from long video sequences, but repeated VLM calls introduce heavy computational overhead.  
Although our \textbf{TLD} strategy greatly reduces the number of candidate latents, each latent still decodes into multiple frames via the 3D VAE, making final-frame selection necessary.

We observe that frames decoded from the last latent generally align well with the editing instructions, making them suitable candidates for final selection. However, due to the inherent characteristics of video diffusion models, motion blur often varies among these frames, and directly choosing a blurred frame can degrade the visual quality.

To address this, we introduce a lightweight \textbf{Self-Consistent Post-Refinement (SCPR)} step. 
Given the decoded frames $\{x_i\}$, we compute a Laplacian-based blur score for each frame,
$s_i = \tfrac{1}{HW}\sum_{u,v} \bigl|\nabla^2 x_i(u,v)\bigr|$~\cite{adrianrosebrockBlurDetectionOpenCV2015,bansalBlurImageDetection2016}, and deterministically select the sharpest frame $x^* = \arg\max_i s_i$. 
Rather than invoking an external deblurring model, we re-inject $x^*$ into the same image-to-video model with a short ``still-video'' prompt (e.g., \emph{``A perfectly still video that enhances image clarity and fine details''}), generating a brief refinement clip. 
We then take the sharpest frame from this clip as the final output $\hat{x}$. 
By leveraging the model’s own temporal prior, SCPR provides a self-consistent post-diffusion enhancement that improves texture fidelity and visual clarity with only a small cost.

\input{figure_latex/tedbench_results_qualitative}

%% file: figure_latex/tedbench_results_qualitative.tex
\begin{figure}[t]
\centering
\scriptsize
\includegraphics[width=\linewidth]{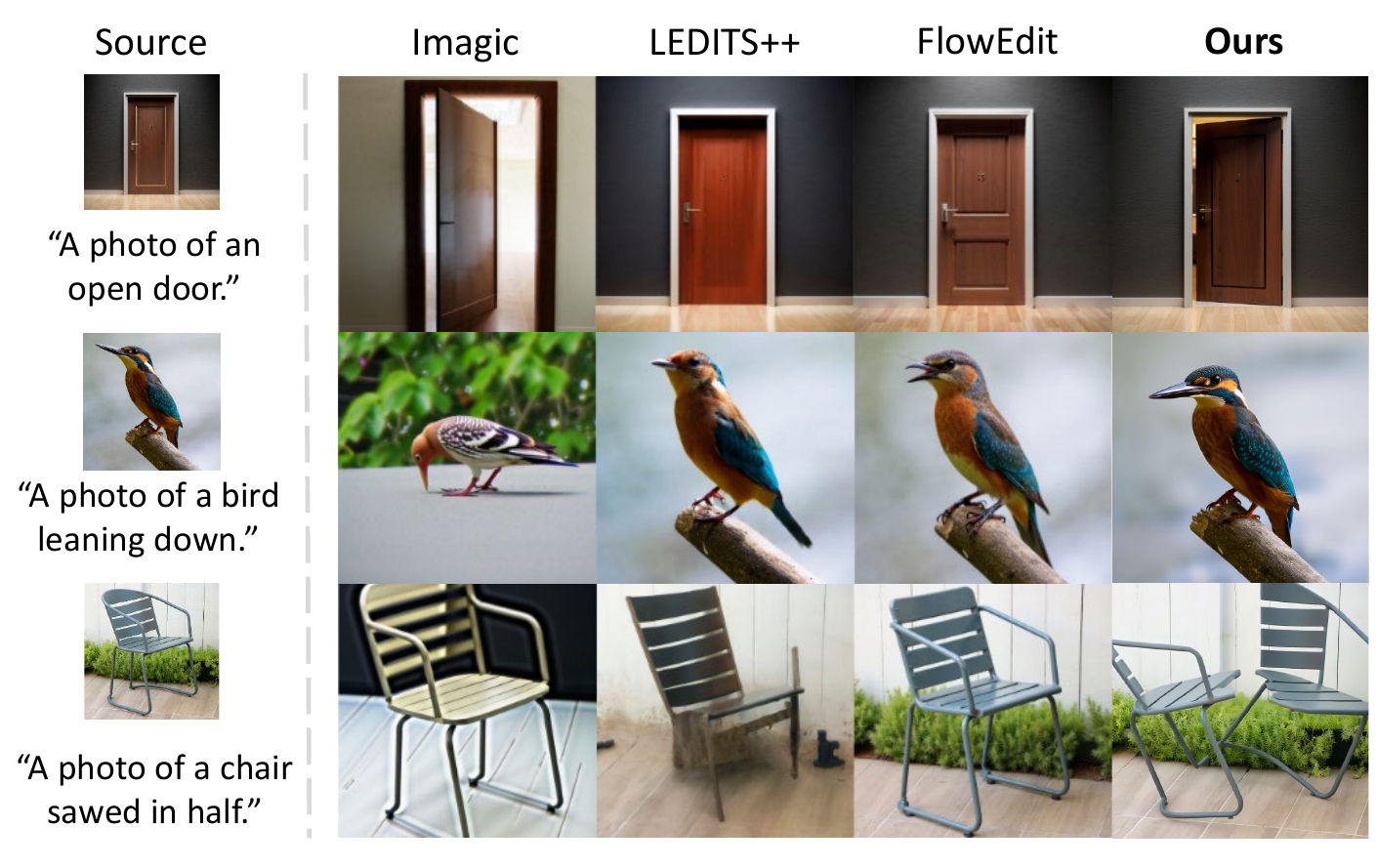}  
 \vspace{-2.0 em}
\scriptsize
\caption{Qualitative Results on TEdBench (\S\ref{sec:non-rigid evaluation}). Comparison with other methods across non-rigid editing tasks.}
\vspace{-1.0 em}
\label{fig:tedbench results qualitative}
\end{figure} 

%% file: sec/4_Experiment.tex
\input{table/TedBench}

\section{Experiment}\label{sec:experiment}

\noindent\textbf{Implementation Details.}
We use Qwen3-VL-30B-A3B-Instruct~\cite{qwen3vl2024} as our prompt enhancer with a customized system instruction for temporal reasoning.  
For generation, we adopt Wan2.2-A14B I2V with Lightning-LoRA~\cite{lightx2v} for accelerated inference.  
We generate 32-frame videos with 8 denoising steps, a dropout threshold of 0.9 (from $t{=}1\!\to\!0$), and a temporal stride of $K{=}3$.  
Each editing run takes \(12\) seconds on a single NVIDIA H100 80GB GPU.

\noindent\textbf{Evaluation Protocol.}  
We assess video-model-based image editing from three complementary perspectives:  
(1) \textbf{Non-Rigid and Motion Editing} using \textbf{TEdBench}~\cite{Kawar_2023_CVPR} and \textbf{ByteMorph}~\cite{chang2025bytemorphbenchmarkinginstructionguidedimage}, following their original metrics;  
(2) \textbf{Reasoning Ability} on \textbf{RISEBench}~\cite{zhao2025envisioningpixelsbenchmarkingreasoninginformed}, covering temporal, causal, spatial, and logical reasoning with GPT-4.1 scoring~\cite{openai2025gpt41};  
(3) \textbf{General Editing}—addition, removal, replacement, style—on \textbf{ImgEdit}~\cite{ye2025imgedit}.  
All experiments use fixed seeds, and we report single-run results for comparability.

\input{table/ByteMorph}
\input{figure_latex/bytemorph_results_qualitative}

\input{figure_latex/risebench_figure}
\input{figure_latex/risebench_results}

\subsection{Evaluation Results on Different Benchmarks}\label{sec: evaluation results}

\noindent\textbf{Non-Rigid and Motion Editing.}\label{sec:non-rigid evaluation}
We evaluate \ifedit{IF-Edit} on non-rigid and motion-centric editing tasks, where temporal reasoning is crucial.  
Our method surpasses both training-free baselines~\cite{mengSDEditGuidedImage2022,parmarZeroshotImagetoImageTranslation2023,Kawar_2023_CVPR,Brack_2024_CVPR,rotstein2025pathwaysimagemanifoldimage,kulikovFlowEditInversionFreeTextBased2025} and fine-tuned models~\cite{brooks2023instructpix2pix,zhang2023magicbrush,zhao2024ultraedit,yu2025anyedit,liu2025step1x,caiHiDreamI1HighEfficientImage2025}, confirming the intrinsic advantage of video diffusion models for such edits.  
On \textbf{TEdBench} (\cref{tab:tedbench,fig:tedbench results qualitative}), which involves deformation and motion scenarios (e.g., a bird spreading wings, a door opening), \ifedit{IF-Edit} achieves the highest CLIP-T and a CLIP-I of 0.96, markedly higher than the previous best of 0.89, showing strong text alignment and image consistency.  

On \textbf{ByteMorph} (\cref{tab:bytemorph_results,fig:bytemorph results qualitative}), covering over 600 samples of camera motion, object interaction, and human actions, we follow the official VLM-based evaluation for semantic accuracy and realism.  
\ifedit{IF-Edit} excels on camera-related edits—especially \emph{Camera Zoom} and \emph{Camera Move}—and maintains strong performance on \emph{Human Motion} and \emph{Interaction}, demonstrating robust temporal reasoning and non-rigid transformation ability.  
These results highlight that video diffusion models naturally capture motion dynamics and world-consistent changes, making them particularly effective for non-rigid editing tasks.

\input{table/RISEBench}

\noindent\textbf{Reasoning Ability.}\label{sec:reasoning evaluation}
We further evaluate \ifedit{IF-Edit} on RISEBench (\cref{tab:risebench_overall,fig:rise-score,fig:risebench results qualitative}), which tests four reasoning types—\emph{Temporal}, \emph{Causal}, \emph{Spatial}, and \emph{Logical}—alongside sub-dimensions reflecting overall reasoning faithfulness and visual quality. 
Compared with open-source baselines~\cite{wu2025qwen,dengEmergingPropertiesUnified2025,labsFLUX1KontextFlow2025,wang2025ovisu1technicalreport,liu2025step1x,xiao2025omnigen,caiHiDreamI1HighEfficientImage2025}, \ifedit{IF-Edit} demonstrates clear advantages in \emph{Temporal} and \emph{Causal} reasoning, benefiting from the video diffusion prior that models frame-to-frame evolution. 
Our Chain-of-Thought prompt enhancement~\cite{wei2022chain} further boosts instruction understanding, producing temporally coherent transformations that align with causal context. 
Although some baselines~\cite{wang2025ovisu1technicalreport,wu2025qwen,dengEmergingPropertiesUnified2025,liu2025step1x} also integrate VLM modules, \ifedit{IF-Edit} achieves higher overall reasoning accuracy under identical evaluation. 

As shown in \cref{fig:rise-score}, \ifedit{IF-Edit} maintains strong \emph{Appearance Consistency} and \emph{Visual Plausibility}, rivaling large closed-source models, while slightly lagging in \emph{Instruction Reasoning}.
This reflects a key trade-off—our method emphasizes physical-temporal realism and coherence over linguistic abstraction—highlighting both the strength and current boundary of video-based editing for reasoning-rich tasks.

\noindent\textbf{General Editing.}\label{sec: general edit evaluation}
We further evaluate \ifedit{IF-Edit} on the \textbf{ImgEdit} benchmark (\cref{tab:imgedit}), which covers diverse instruction-based edits such as object addition, removal, replacement, and style transfer, with GPT-4.1 as the evaluator. 
As a training-free video-based method, \ifedit{IF-Edit} performs competitively with early instruction-tuned models~\cite{dai2023instructblip,zhang2023magicbrush,yu2025anyedit}, indicating strong generalization despite the absence of paired supervision. 
Notably, in the \emph{Action} category—where dynamic realism and structural coherence are crucial—our method surpasses most conventional systems, and qualitatively better preserves global pose and layout under large, instruction-driven changes.

Compared with recent large-scale image editing systems~\cite{liu2025step1x,dengEmergingPropertiesUnified2025,zhang2025context,linUniWorldV1HighResolutionSemantic2025,wu2025omnigen2,zhao2024ultraedit}, however, \ifedit{IF-Edit} still lags behind in overall accuracy. 
Without fine-tuning for region-specific manipulation, it sometimes interprets edits as global scene updates, leading to viewpoint shifts or imperfect preservation of unedited areas, especially for small attribute or style changes. 
This behavior reflects the inductive bias of image-to-video models toward modeling holistic dynamics rather than strictly localized appearance edits. 
Even so, its strong structural coherence and realistic rendering suggest that video priors are a promising basis for unified, tuning-free editing, and could be further enhanced with region-aware controls or targeted fine-tuning.


\subsection{Ablation Study}\label{sec:ablation}
We evaluate the contribution of each component of \ifedit{IF-Edit} on TEdBench (\cref{tab:ablation}). 
Ablating Chain-of-Thought prompt enhancement (\textit{no-enhance}) noticeably lowers CLIP-T scores (0.65$\rightarrow$0.59), indicating that temporally grounded reasoning prompts are important for aligning the video model’s world-simulation prior with the target edit. 
Disabling Temporal Latent Dropout (\textit{K=1}) nearly doubles inference time (12s$\rightarrow$21s) with only marginal quality gain, suggesting that most temporal latents are redundant once the global layout has been established; among sparsification intervals, \textit{K=3} offers the best trade-off, while overly aggressive dropout (\textit{K=4}) starts to harm temporal consistency. 
Removing the refinement step (\textit{no-refine}) reduces sharpness (983$\rightarrow$840), showing that the brief still-video trajectory helps recover fine details without altering semantics. 
Finally, replacing our lightweight refinement with a VLM-based filter achieves comparable accuracy but increases runtime substantially (12s$\rightarrow$37s), highlighting that self-consistent refinement provides a more practical balance between quality and efficiency.

\input{table/ImgEdit}

\subsection{Further Discussion}

\input{table/ablation_table}

\noindent\textbf{Application.}\label{sec:application}
Video diffusion models extend image editing into a physically grounded and temporally coherent paradigm.
Their strong world-simulation priors enable natural manipulations such as motion, deformation, and environmental change while maintaining global consistency (see~\cref{fig:tedbench results qualitative,fig:bytemorph results qualitative}).
Moreover, visualizing their temporal reasoning process reveals how the model “thinks” through causal and physical transformations (see~\cref{fig:teaser}), offering insights into generative reasoning.
These trajectories could further serve as supervision for distilling video priors into compact image reasoning models or training MLLMs and world models with explicit temporal structure.

\input{figure_latex/imgedit_failed}

\noindent\textbf{Limitation.}\label{sec:limitation}
While \ifedit{IF-Edit} shows strong editing performance, two limitations remain.
\textbf{(1) Limited general instruction editing.}
Without task-specific finetuning, performance drops on region-based or highly abstract edits.
Video diffusion priors favor physically plausible, temporally smooth transformations—effective for non-rigid and reasoning-like edits—but struggle with unrealistic insertions or replacements (see~\cref{fig:imgedit-failed}).
Finetuning or guided region-aware control may alleviate this.
\textbf{(2) High GPU memory usage.}
Despite the acceleration from Temporal Latent Dropout, video backbones still demand large memory ($>40$~GB) due to multi-frame processing.
Model compression, quantization, or pruning~\cite{changSparseDiTTokenSparsification2025, chenQDiTAccuratePostTraining2025, ganjdaneshMixtureEfficientDiffusion2025, gaoModulatedDiffusionAccelerating2025, heEfficientDMEfficientQuantizationAware2023, jiaD^2iTDynamicDiffusion2025, leeTextEmbeddingKnows2025, liSVDQuantAbsorbingOutliers2024, liuCacheQuantComprehensivelyAccelerated2025, ryuDGQDistributionAwareGroup2024, shaoMemoryEfficientGenerativeModels2025, suiBitsFusion199Bits2024, wangQuESTLowbitDiffusion2025, youLayerTimestepAdaptiveDifferentiable2025, zhaoDynamicDiffusionTransformer2024, zhuSimpleDistillationOneStep2025} could reduce cost, though at the expense of speed.

%% file: table/TedBench.tex
\begin{table}[t]
	\centering
	\caption{\textbf{Quantitative results on TEdBench. (\S\ref{sec:non-rigid evaluation})} Following~\cite{rotstein2025pathwaysimagemanifoldimage}, CLIP-I~\cite{radford2021learning} and LPIPS~\cite{lpips} are calculated between source and edited image and CLIP-T~\cite{radford2021learning} is calculated between edited image and target prompt.  }\label{tab:tedbench}
	\vspace{-3.5mm}
\centering
\renewcommand\arraystretch{1.1}
\resizebox{\linewidth}{!}{
\begin{tabular}{rcccc}
\bottomrule[1pt]\rowcolor[HTML]{FAFAFA}

\multicolumn{1}{c}{Methods} & Venue  & $\text{LPIPS}\downarrow$         & $\text{CLIP-I}\uparrow$ & $\text{CLIP-T}\uparrow$ \\ 
\toprule[0.8pt]

\rowcolor[HTML]{F9FFF9}
SDEdit~\cite{mengSDEditGuidedImage2022}  & ICLR22 & 0.30     & 0.85 & 0.60           \\ 

\rowcolor[HTML]{F9FFF9}
Pix2Pix-ZERO~\cite{parmarZeroshotImagetoImageTranslation2023}  & SIGGRAPH23 & 0.29       & 0.84 & 0.62         \\ 

\rowcolor[HTML]{F9FFF9} Imagic~\cite{Kawar_2023_CVPR}  & CVPR23 & 0.52     & 0.86 & 0.63   \\ 
\rowcolor[HTML]{F9FFF9} LEDITS++~\cite{Brack_2024_CVPR}  & CVPR24 &  0.23  & 0.87 &0.63  \\
\rowcolor[HTML]{F9FFF9} F2F~\cite{rotstein2025pathwaysimagemanifoldimage}  & CVPR25 & 0.22  & 0.89 & 0.63           \\
\rowcolor[HTML]{F9FFF9} FlowEdit~\cite{kulikovFlowEditInversionFreeTextBased2025}  & ICCV25 & 0.22  &  0.89 & 0.61 \\

\toprule[0.8pt]

\rowcolor[HTML]{F9FBFF} \multicolumn{1}{r}{\ \ \ \
 \textbf{\ifedit{IF-Edit}} (ours)} & - &  \textbf{0.19}  & \textbf{0.96} & \textbf{0.65}   \\

\toprule[1pt]
\end{tabular}}
\vspace{-4.0mm}
\end{table}

%% file: table/ByteMorph.tex
\begin{table}[t]
  \centering
  \caption{\textbf{Results on ByteMorph. (\S\ref{sec:non-rigid evaluation})} We report VLM-evaluation score (based on Claude-3.7-Sonnet) on \textit{Camera Zoom}, \textit{Camera Move}, \textit{Object Motion}, \textit{Human Motion}, and \textit{Interaction}.}
  \label{tab:bytemorph_results}
  \vspace{-3.5mm}
\renewcommand\arraystretch{1.1}
\resizebox{\linewidth}{!}{
\begin{tabular}{rccccc}
\bottomrule[1pt]\rowcolor[HTML]{FAFAFA}
\multicolumn{1}{c}{Method} &
\makecell{Camera\\Zoom} &
\makecell{Camera\\Move} &
\makecell{Object\\Motion} &
\makecell{Human\\Motion} &
\makecell{Inter-\\action} \\
\toprule[0.8pt]

\rowcolor[HTML]{F9FFF9}
InstructPix2Pix~\cite{brooks2023instructpix2pix} & 42.37 & 32.20 & 36.47 & 23.60 & 31.29 \\

\rowcolor[HTML]{F9FFF9}
MagicBrush~\cite{zhang2023magicbrush}      & 49.37 & 52.63 & 47.49 & 46.27 & 39.98 \\

\rowcolor[HTML]{F9FFF9}
UltraEdit (SD3)~\cite{zhao2024ultraedit} & 54.74 & \underline{59.01} & 62.13 & 50.64 & 52.24 \\

\rowcolor[HTML]{F9FFF9}
AnyEdit~\cite{yu2025anyedit}          & 40.92 & 49.37 & 48.31 & 38.12 & 37.23\\

\rowcolor[HTML]{F9FFF9}
Step1X-Edit~\cite{liu2025step1x}      & \underline{59.34} & 57.96 & \textbf{72.78} & \underline{65.39} & \underline{65.99} \\

\rowcolor[HTML]{F9FFF9}
HiDream-E1~\cite{caiHiDreamI1HighEfficientImage2025} & 41.18 & 32.76 & 35.00& 33.21 & 35.73 \\

\toprule[0.8pt]

\rowcolor[HTML]{F9FBFF} \multicolumn{1}{r}{\ \ \ \
 \textbf{\ifedit{IF-Edit}} (ours)} & \textbf{67.89} & \textbf{59.80} & \underline{72.02} & \textbf{67.04} & \textbf{69.05} \\

\toprule[1pt]
\end{tabular}}
\vspace{-3.0mm}
\end{table}

%% file: figure_latex/bytemorph_results_qualitative.tex
\begin{figure}[t]
\centering
\scriptsize
\includegraphics[width=\linewidth]{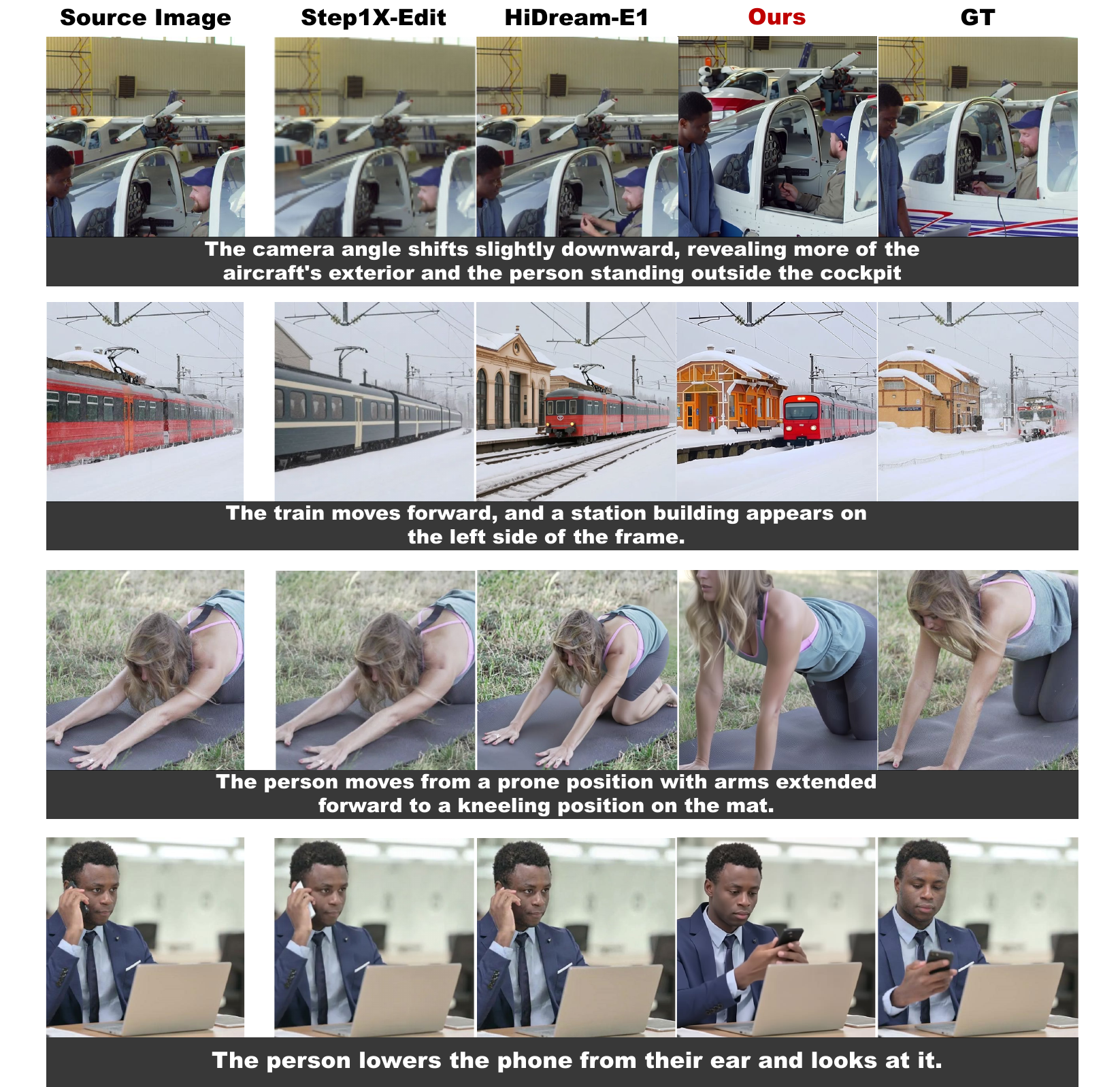}  
 \vspace{-2.0 em}
\scriptsize
\caption{Qualitative Results on ByteMorph Benchmark (\S\ref{sec:non-rigid evaluation}). Comparison with other methods on non-rigid editing tasks. The editing types from top to bottom are camera motion, object motion, human motion and human interaction.}
\vspace{-2.0 em}
\label{fig:bytemorph results qualitative}
\end{figure}

%% file: figure_latex/risebench_figure.tex
\begin{figure}[t]
\centering
\scriptsize
\includegraphics[width=\linewidth]{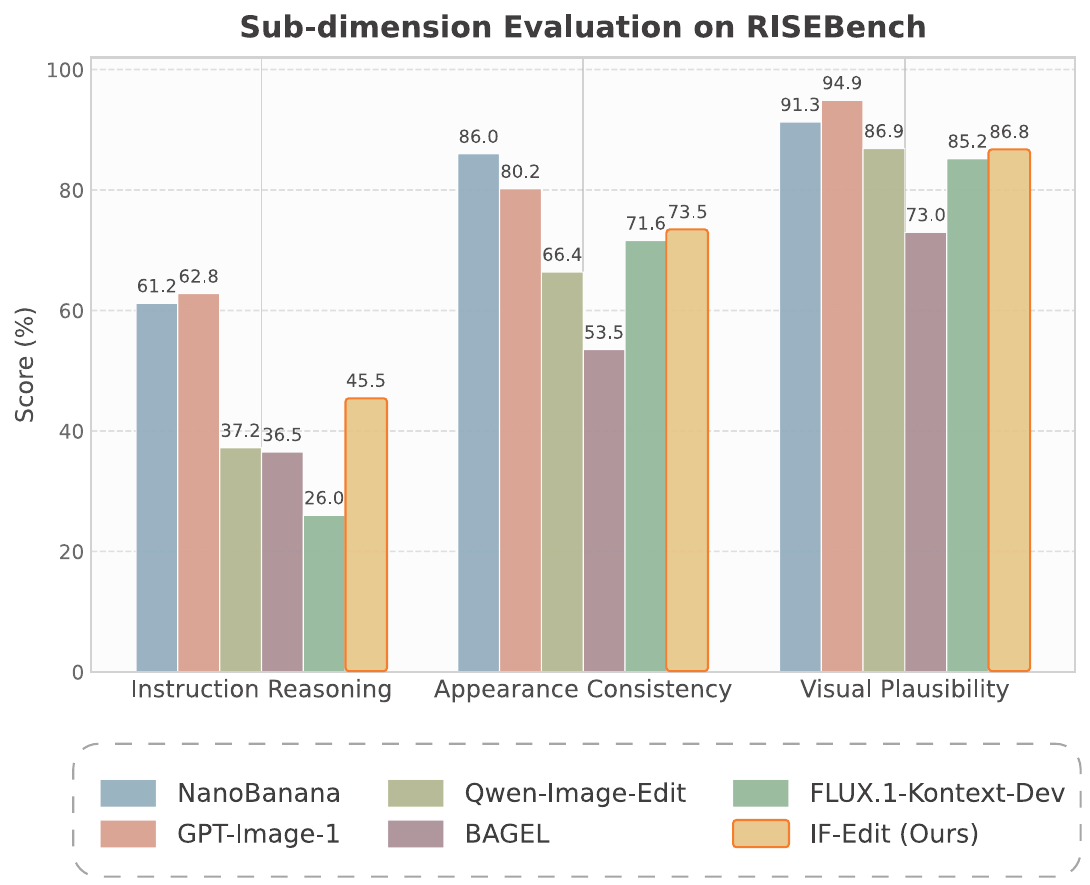}  
 \vspace{-2.0 em}
\scriptsize
\caption{Comparison across models on three evaluation sub-dimensions in RISEBench (\S\ref{sec:reasoning evaluation}). Results evaluated by GPT-4.1.}
\vspace{-1.0 em}
\label{fig:rise-score}
\end{figure} 

%% file: figure_latex/risebench_results.tex
\newcommand{\figureCaption}{
\textbf{Results on RISEBench (\S\ref{sec:reasoning evaluation}).}
\ifedit{IF-Edit} produces more natural and coherent transformations than conventional models, with edits that better respect physical laws, temporal progression, and spatial structure thanks to the video model’s world-consistent priors.
}

\begin{figure*}[tbp]
    \centering
    \scriptsize
    \includegraphics[width=\linewidth]{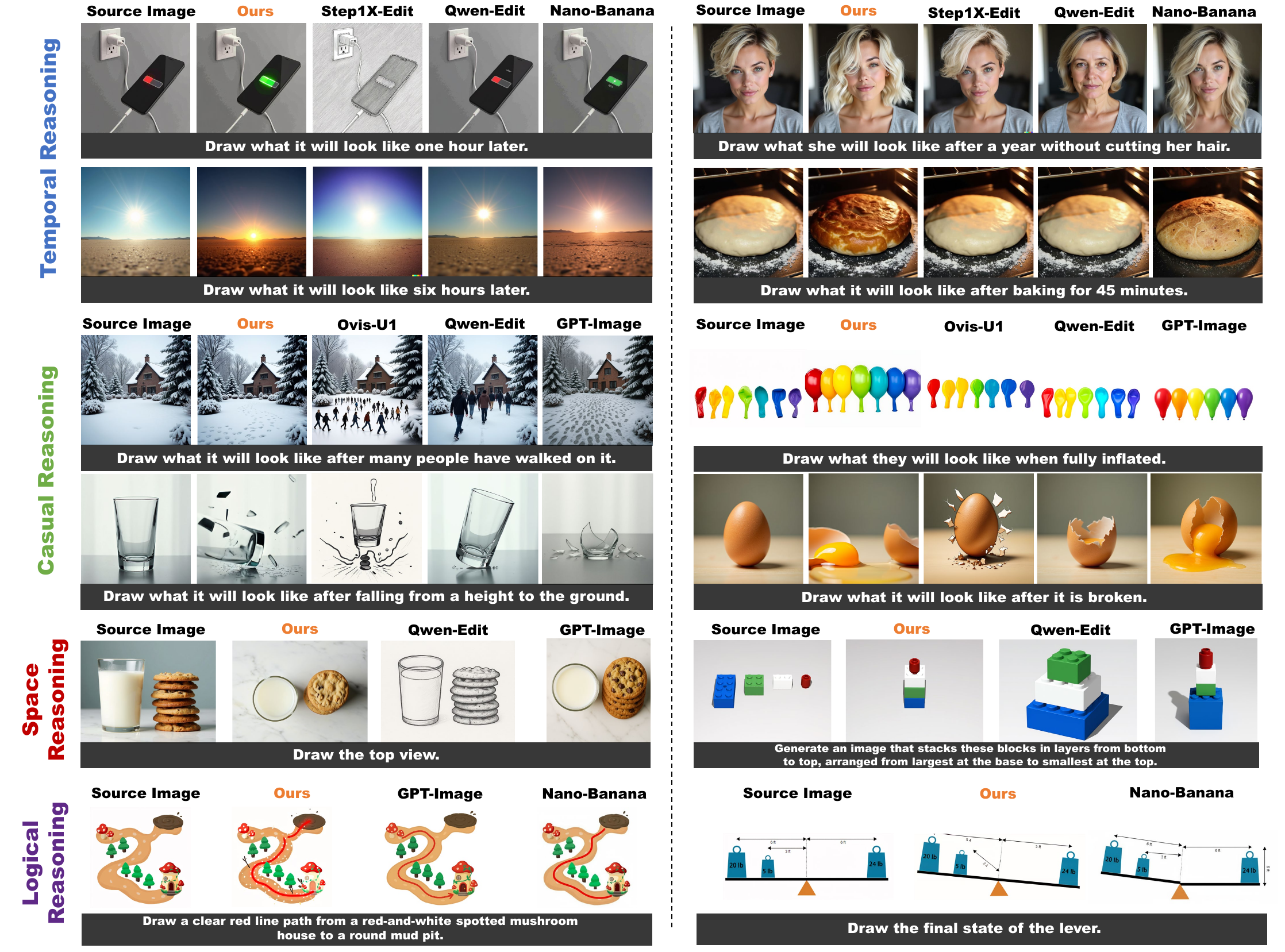}
    \vspace{-1.5 em}
    \captionof{figure}{\figureCaption}
     \vspace{-1.5 em}
    \label{fig:risebench results qualitative}
\end{figure*}

%% file: table/RISEBench.tex
\begin{table}[t]
  \centering
  \caption{\textbf{Overall performance on RISEBench. (\S\ref{sec:reasoning evaluation})} We report category-wise accuracy (\%) evaluated by GPT-4.1 on \textit{Temporal}, \textit{Causal}, \textit{Spatial}, \textit{Logical}, and the \textit{Overall} average. GPT-Image and Nano-Banana are comercial models.}
  \label{tab:risebench_overall}
  \vspace{-3.5mm}
\large
\renewcommand\arraystretch{1.1}
\resizebox{\linewidth}{!}{
\begin{tabular}{rccccc}
\bottomrule[1pt]\rowcolor[HTML]{FAFAFA}
\multicolumn{1}{c}{Model} & Temporal (\%) & Causal (\%) & Spatial (\%) & Logical (\%) & Overall (\%) \\
\toprule[0.8pt]

\rowcolor[HTML]{F9FBFF}
Nano-Banana~\cite{google2025gemini25flashmodelcard}     & 25.9 & 47.8 & 37.0 & 18.8 & 32.8 \\
\rowcolor[HTML]{F9FBFF}
GPT-Image-1~\cite{gpt4o}                & 34.1   & 32.2          & 37.0 & 10.6 & 28.9 \\

\toprule[0.1pt]

\rowcolor[HTML]{F9FFF9}
Qwen-Image-Edit~\cite{wu2025qwen}            & \underline{4.7}              & \underline{10.0}          & \textbf{17.0}          & 2.4            & \underline{8.9} \\

\rowcolor[HTML]{F9FFF9}
BAGEL~\cite{dengEmergingPropertiesUnified2025}                      & 2.4              & 5.6           & \underline{14.0}          & 1.2            & 6.1 \\

\rowcolor[HTML]{F9FFF9}
Kontext-Dev~\cite{labsFLUX1KontextFlow2025}         & 2.3              & 5.5           & 13.0          & 1.2            & 5.8 \\

\rowcolor[HTML]{F9FFF9}
Ovis-U1~\cite{wang2025ovisu1technicalreport}                    & 1.2              & 3.3           & 4.0           & 2.4            & 2.8 \\

\rowcolor[HTML]{F9FFF9}
Step1X-Edit~\cite{liu2025step1x}                & 0.0              & 2.2           & 2.0           & \underline{3.5}            & 1.9 \\

\rowcolor[HTML]{F9FFF9}
OmniGen~\cite{xiao2025omnigen}                    & 1.2              & 1.0           & 0.0           & 1.2            & 0.8 \\

\rowcolor[HTML]{F9FFF9}
HiDream-Edit~\cite{caiHiDreamI1HighEfficientImage2025}               & 0.0              & 0.0           & 0.0           & 0.0            & 0.0 \\

\toprule[0.8pt]

\rowcolor[HTML]{F9FBFF} \multicolumn{1}{r}{\ \ \ \
 \textbf{\ifedit{IF-Edit}} (ours)} & \textbf{5.8} &  \textbf{21.1}  & 12.0 & \textbf{4.7} & \textbf{11.1}  \\

\toprule[1pt]
\end{tabular}}
\vspace{-6.0mm}
\end{table}

%% file: table/ImgEdit.tex
\begin{table}[t]
  \centering
  \caption{\textbf{Performance on ImgEdit benchmark (\S\ref{sec: general edit evaluation}).} We report category-wise and overall scores on various editing tasks. We use GPT-4.1 for evaluation following the original paper.}
  \label{tab:imgedit}
  \vspace{-3.5mm}
\large
\renewcommand\arraystretch{1.1}
\resizebox{\linewidth}{!}{
\begin{tabular}{rcccccccccc}
\bottomrule[1pt]\rowcolor[HTML]{FAFAFA}
\multicolumn{1}{c}{Model} & Add & Adjust & Extract & Replace & Remove & BackG & Style & Hybrid & Action & Overall$\uparrow$ \\
\toprule[0.8pt]

\rowcolor[HTML]{F9FFF9}
MagicBrush~\cite{zhang2023magicbrush} & 2.84 & 1.58 & 1.51 & 1.97 & 1.58 & 1.75 & 2.38 & 1.62 & 1.22 & 1.83 \\

\rowcolor[HTML]{F9FFF9}
Instruct-P2P~\cite{brooks2023instructpix2pix} & 2.45 & 1.83 & 1.44 & 2.01 & 1.50 & 1.44 & 3.55 & 1.20 & 1.46 & 1.88 \\

\rowcolor[HTML]{F9FFF9}
AnyEdit~\cite{yu2025anyedit} & 3.18 & 2.95 & 1.88 & 2.47 & 2.23 & 2.24 & 2.85 & 1.56 & 2.65 & 2.45 \\

\rowcolor[HTML]{F9FFF9}
UltraEdit~\cite{zhao2024ultraedit} & 3.44 & 2.81 & 2.13 & 2.96 & 1.45 & 2.83 & 3.76 & 1.91 & 2.98 & 2.70 \\

\rowcolor[HTML]{F9FFF9}
ICEdit~\cite{pan2025ice} & 3.58 & 3.39 & 1.73 & 3.15 & 2.93 & 3.08 & 3.84 & 2.04 & 3.68 & 3.05 \\

\rowcolor[HTML]{F9FFF9}
Step1X-Edit~\cite{liu2025step1x} & 3.88 & 3.14 & 1.76 & 3.40 & 2.41 & 3.16 & 4.63 & 2.64 & 2.52 & 3.06 \\

\rowcolor[HTML]{F9FFF9}
UniWorld-V1~\cite{linUniWorldV1HighResolutionSemantic2025} & 3.82 & 3.64 & 2.27 & 3.47 & 3.24 & 2.99 & 4.21 & 2.96 & 2.74 & 3.26 \\

\rowcolor[HTML]{F9FFF9}
BAGEL~\cite{dengEmergingPropertiesUnified2025} & 3.81 & 3.59 & 1.58 & 3.85 & 3.16 & 3.39 & 4.51 & 2.67 & 4.25 & 3.42 \\

\rowcolor[HTML]{F9FFF9}
OmniGen2~\cite{wu2025omnigen2} & 3.57 & 3.06 & 1.77 & 3.74 & 3.20 & 3.57 & 4.81 & 2.52 & 4.68 & 3.44 \\

\rowcolor[HTML]{F9FFF9}
Kontext-Dev~\cite{labsFLUX1KontextFlow2025} & 3.83 & 3.65 & 2.27 & 4.45 & 3.17 & 3.98 & 4.55 & 3.35 & 4.29 & 3.71 \\

\rowcolor[HTML]{F9FFF9}

\rowcolor[HTML]{F9FFF9}
GPT-Image~\cite{gpt4o} & 4.61 & 4.33 & 2.90 & 4.35 & 3.66 & 4.57 & 4.93 & 3.96 & 4.89 & 4.20 \\

\toprule[0.8pt]

\rowcolor[HTML]{F9FBFF} \multicolumn{1}{r}{\ \ \ \
 \textbf{\ifedit{IF-Edit}} (ours)} & 2.94 & 2.02 & 1.71 & 2.18 & 1.50 & 1.71 & 2.53 & 2.06 & 3.10 & 2.19   \\

\toprule[1pt]
\end{tabular}}

\end{table}

%% file: table/ablation_table.tex
\begin{table}[t]
\centering
\caption{\textbf{Ablation study on TEdBench} (\S\ref{sec:ablation}).
We report CLIP-T (↑), CLIP-I (↑), LPIPS (↓), image sharpness (↑), and inference time (↓). PE: Prompt Enhance, TD(K=x): Temporal Dropout with K=x where TD(K=1) means no dropout, PR: Post-Refine.
Sharpness is measured via Laplacian-based blur score.}
\vspace{-2pt}
\resizebox{\linewidth}{!}{
\begin{tabular}{lccccc}
\toprule
\textbf{Configuration} & \textbf{CLIP-T↑} & \textbf{CLIP-I↑} & \textbf{LPIPS↓} & \textbf{Sharpness↑} & \textbf{Time (s)↓} \\
\midrule
\cellcolor{cyan}w/o Prompt Enhance (\textit{naive prompt}) & 0.59 & 0.95 & 0.20 & 981 & 10 \\
\cellcolor{cyan}w/o Post-Refine (\textit{no-refine})      & 0.63 & 0.94 & 0.23 & 840 & \textbf{7} \\
\cellcolor{cyan}PE + PR + TD(K=x) & --- & --- & --- & --- & --- \\
\ \ \ \ K=1       & \textbf{0.65} & \textbf{0.96} & \textbf{0.17} & \textbf{983} & 21 \\
\ \ \ \ K=2                                   & 0.64 & 0.95 & 0.20 & 982 & 15 \\
\cellcolor{ifeditlight}\ \ \ \ K=3 (\textit{Ours})           & \textbf{0.65} & \textbf{0.96} & 0.19 & \textbf{983} & 12 \\
\ \ \ \ K=4                                   & 0.62 & 0.92 & 0.22 & 927 & 11 \\
\midrule
\cellcolor{pink}w/ VLM filter                   & 0.64   &  0.95    &  0.21  &   895   &  37\\
\bottomrule
\end{tabular}
}
\label{tab:ablation}
\vspace{-1.5em}
\end{table}

%% file: figure_latex/imgedit_failed.tex
\begin{figure}[t]
\centering
\scriptsize
\adjustbox{width=.9\linewidth, trim={0.1\width} {0.00\height} {0.1\width} {0.02\height}, clip}{
  \includegraphics{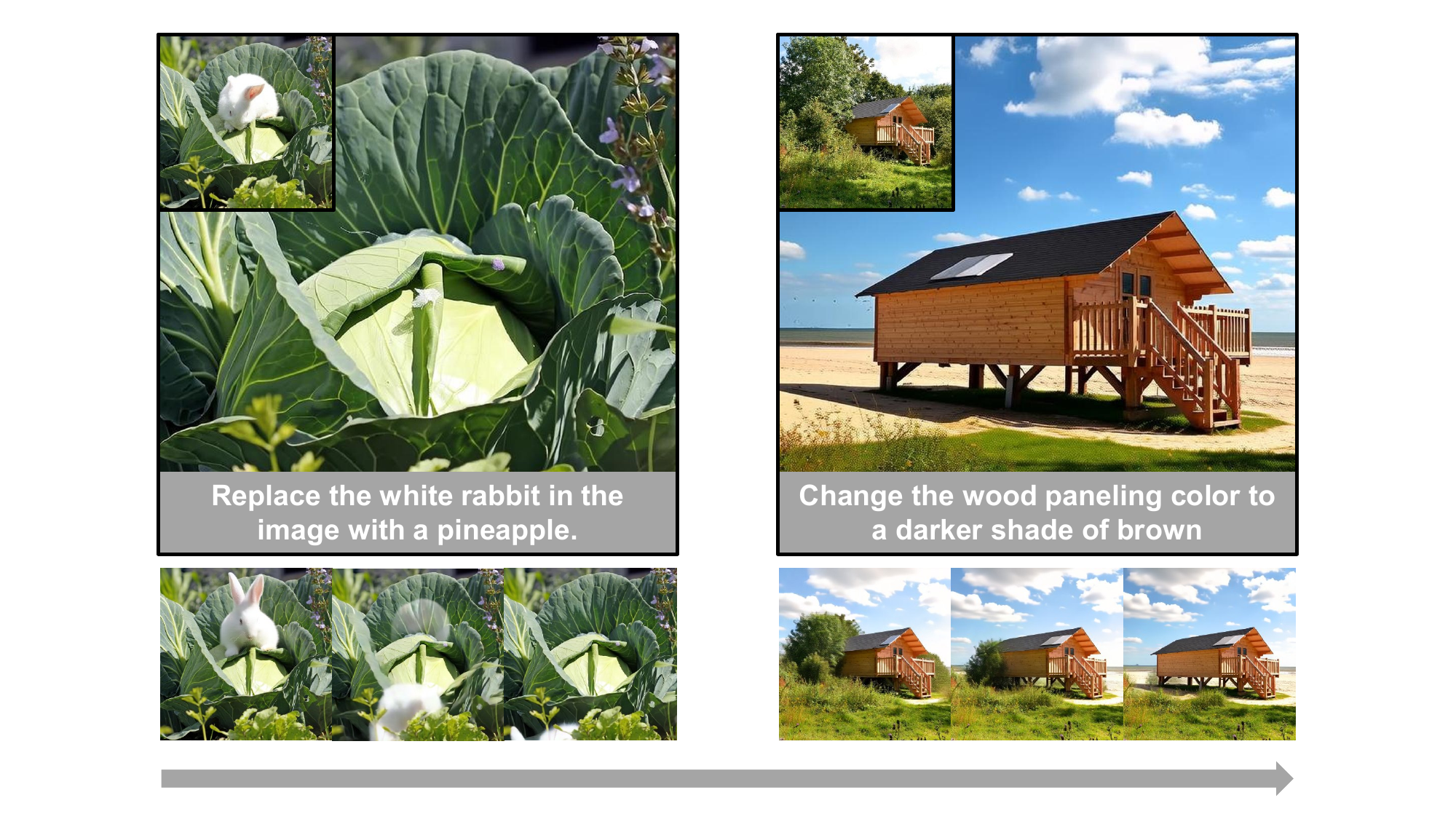}  
}
 \vspace{-2.0 em}
\scriptsize
\caption{Failure cases (\S\ref{sec:limitation}).  
Video models keep strong global consistency but struggle with large object-level modifications.}

 \vspace{-6.0 mm}
\label{fig:imgedit-failed}
\end{figure}

%% file: sec/5_Conclusion.tex
\section{Conclusion}\label{sec: conclusion}
We introduced \ifedit{IF-Edit}, a tuning-free framework that adapts image-to-video diffusion models for zero-shot image editing. 
Our method integrates three lightweight modules—Prompt Enhancement via Chain-of-Thought, Temporal Latent Dropout, and Self-Consistent Post-Refinement—to enhance reasoning, improve efficiency, and refine visual fidelity. 
For the first time, we systematically evaluate video diffusion models across multiple image-editing benchmarks, revealing that they excel in \textbf{non-rigid} and \textbf{reasoning-centric} edits while remaining competitive in general instruction-based tasks. 
Although performance on region-specific edits still trails specialized image editors, our findings highlight the untapped potential of video diffusion as a foundation for unified image editing and visual reasoning. 
We hope this work inspires future research toward fine-tuned or hybrid video–image editors and broader applications in visual world modeling.